# Energy Distribution of EEG Signals: EEG Signal Wavelet-Neural Network Classifier

I. Omerhodzic, S. Avdakovic, A. Nuhanovic, K. Dizdarevic

*Abstract*— In this paper, a wavelet-based neural network (WNN) classifier for recognizing EEG signals is implemented and tested under three sets EEG signals (healthy subjects, patients with epilepsy and patients with epileptic syndrome during the seizure). First, the Discrete Wavelet Transform (DWT) with the Multi-Resolution Analysis (MRA) is applied to decompose EEG signal at resolution levels of the components of the EEG signal (δ, θ, α, β and γ) and the Parseval's theorem are employed to extract the percentage distribution of energy features of the EEG signal at different resolution levels. Second, the neural network (NN) classifies these extracted features to identify the EEGs type according to the percentage distribution of energy features. The performance of the proposed algorithm has been evaluated using in total 300 EEG signals. The results showed that the proposed classifier has the ability of recognizing and classifying EEG signals efficiently.

## I. Introduction

EPILEPSY is one of the world's most common neurological diseases, affecting more than 40 million people worldwide. Epilepsy's hallmark symptom, seizures, can have a broad spectrum of debilitating medical and social consequences [1]. Although antiepileptic drugs have helped treat millions of patients, roughly a third of all patients are unresponsive to pharmacological intervention. Understanding of this dynamic disease evolves; new possibilities for treatment are emerging. An area of great interest is the development of devices that incorporate algorithms capable of detecting early onset of seizures or even predicting those hours before they occur. This lead time will allow for new types of interventional treatment. In the near future a patient's seizure may be detected and aborted before physical manifestations begin.

Ibrahim Omerhodzic, MD, MSc, Department of neurosurgery, Clinical Center University of Sarajevo, Bolnicka 25, 71000 Sarajevo, Bosnia and Herzegovina (corresponding author) phone: +387 33 29 81 06; fax: +387 33 22 09 94; (e-mail: genijana@bih.net.ba).
Samir Avdakovic, MSc, El. Eng., EPC Elektroprivreda BiH D.D. Sarajevo, Department for Development, Vilsonovo setaliste 15, 71000 Sarajevo, Bosnia and Herzegovina (e-mail: s.avdakovic@elektroprivreda.ba).
Amir Nuhanovic, PhD, El. Eng., University of Tuzla, Faculty of Electrical Engineering, Department of Power Systems Analysis, Franjevacka 2, 75000 Tuzla, Bosnia and Herzegovina (e-mail: amir.nuhanovic@untz.ba).
Kemal Dizdarevic, MD, PhD, Department of neurosurgery, Clinical Center University of Sarajevo, Bolnicka 25, 71000 Sarajevo, Bosnia and Herzegovina (e-mail: kemaldiz@bih.net.ba).

Electroencephalogram (EEG) has established itself as an important means of identifying and analyzing epileptic seizure activity in humans. In most cases, identification of the epileptic EEG signal is done manually by skilled professionals, who are small in number [2]. The diagnosis of an abnormal activity of the brain functionality is a vital issue. EEG signals involve a great deal of information about the function of the brain. But classification and evaluation of these signals are limited. Since there is no definite criterion evaluated by the experts, visual analysis of EEG signals in time domain may be insufficient. Routine clinical diagnosis needs to analysis of EEG signals. Therefore, some automation and computer techniques have been used for this aim.

Recent applications of the wavelet transform (WT) and neural network (NN) to engineering-medical problems can be found in several studies that refer primarily on the signal processing and classification in different medical area. WT applied for EEG signal analyses and WNN applied for classification of EEG signals is not a new concept. Several papers in different ways applied WT to analyze EEG signals and combine the WT and NN in the process of classification. Some of the papers are listed in the references [2]-[27].

This paper presents an algorithm for classification of EEG signals based on wavelet transformation (WT) and patterns recognize techniques. Discrete Wavelet Transform (DWT) with the Multi-Resolution Analysis (MRA) is applied to decompose EEG signal at resolution levels of the components of the EEG signal (δ, θ, α, β and γ) and the Parseval's theorem are employed to extract the percentage distribution of energy features of the EEG signal at different resolution levels. The neural network (NN) classifies these extracted features to identify the EEGs type according to the percentage distribution of energy features.

The paper is organized as follow. The methodology the proposed process is presented in Section II of this paper. Test results of classifications are given in Section III. Conclusions are given in Section IV.

## II. Methodology

### A. Wavelet Transform

The wavelet transform (WT) introduces a useful representation of a function in the time-frequency domain [28-31]. Basically, a wavelet is a function $\psi \in L^2(R)$ with a zero average

$$\int_{-\infty}^{+\infty} \psi(t)dt = 0. \qquad (1)$$

The Continuous Wavelet Transformation (CWT) of a signal $x(t)$ is then defined as:

$$CWT_\psi x(a,b) = \frac{1}{\sqrt{|a|}} \int_{-\infty}^{+\infty} x(t)\psi^*\left(\frac{t-b}{a}\right)dt \qquad (2)$$

where $\psi(t)$ is called the mother wavelet, the asterisk denotes complex conjugate, while $a$ and $b$ $(a, b \in R)$ are scaling (dilation and translation) parameters, respectively. The scale parameter $a$ determines the oscillatory frequency and the length of the wavelet, and the translation parameter $b$ determines its shifting position.

The application of WT in engineering areas usually requires the discrete WT (DWT). The DWT is defined by using discrete values of the scaling parameter $a$ and the translation parameter $b$. To do so, set $a = a_0^m$ and $b = nb_0 a_0^m$, then we get $\psi_{m,n}(t) = a_0^{-m/2} \psi(a_0^{-m} t - nb_0)$, where $m, n \in Z$, and $m$ is indicating frequency localization and $n$ is indicating time localization. Generally, we can choose $a_0 = 2$ and $b_0 = 1$. This choice will define a dyadic-orthonormal WT and provide the basis for multi-resolution analysis (MRA). In MRA, any time series $x(t)$ can be completely decomposed in terms of approximations, provided by scaling functions $\phi_m(t)$ (also called father wavelet) and the details, provided by the wavelets $\psi_m(t)$. The scaling function is associated with the low-pass filters (LPF), and the wavelet function is associated with the high-pass filters (HPF). The decomposition procedure starts by passing a signal through these filters. The approximations are the low-frequency components of the time series and the details are the high-frequency components. The signal is passed through a HPF and a LPF. Then, the outputs from both filters are decimated by 2 to obtain the detail coefficients and the approximation coefficients at level 1 (A1 and D1). The approximation coefficients are then sent to the second stage to repeat the procedure. Finally, the signal is decomposed at the expected level.

According to Parseval's theorem, the energy of the distorted signal can be partitioned at different resolution levels. Mathematically this can be presented as:

$$ED_i = \sum_{j=1}^{N} |D_{ij}|^2, \quad i = 1,\ldots,l \qquad (3)$$

$$EA_l = \sum_{j=1}^{N} |A_{lj}|^2 \qquad (4)$$

where $i = 1,\ldots,l$ is the wavelet decomposition level from level 1 to level $l$. N is the number of the coefficients of detail or approximate at each decomposition level. $ED_i$ is the energy of the detail at decomposition level $i$ and $EA_l$ is the energy of the approximate at decomposition level $l$.

### B. Artificial Neural Networks

Artificial neural networks (ANNs) are formed of cells simulating the low-level functions of biological neurons. In ANN, knowledge about the problem is distributed in neurons and connections weights of links between neurons [15]. The neural network must be trained to adjust the connection weights and biases in order to produce the desired mapping. At the training stage, the feature vectors are applied as input to the network and the network adjusts its variable parameters, the weights and biases, to capture the relationship between the input patterns and outputs. ANNs are particularly useful for complex pattern recognition and classification tasks. ANNs are widely used in the biomedical field for modeling, data analysis and diagnostic classification. The most frequently used training algorithm in classification problems is the back-propagation (BP) algorithm, which is used in this work also.

There are many different types and architectures of neural networks varying fundamentally in the way they learn, the details of which are well documented in the literature [34-36].

### C. Proposed Methodology

The clinical interests in (EEG) are; for example, sleep pattern analysis, cognitive tasks registration, seizure and epilepsy detection, and other states of the brain, both normal and patho-physiological. Epilepsy is the second most prevalent neurological disorder in humans after stroke. It is characterized by recurring seizures in which abnormal electrical activity in the brain causes altered perception or behavior. Well-known causes of epilepsy may include: genetic disorders, traumatic brain injury, metabolic disturbances, alcohol or drug abuse, brain tumor, stroke, infection, and cortical malformations (dysplasia).

The EEG signal contains a several spectral components. The amplitude of a human surface EEG signal is in the range of 10 to 100 μV. The frequency range of the EEG has a fuzzy lower and upper limit, but the most important frequencies from the physiological viewpoint lie in the range of 0.1 to 30 Hz. The standard EEG clinical bands are the delta (0.1 to 3.5 Hz), theta (4 to 7.5 Hz), alpha (8 to 13 Hz), and beta (14 to 30 Hz) bands [1]. EEG signals with frequencies greater than 30 Hz are called gamma waves.

The datasets used in this research are selected from the Epilepsy center in Bonn, Germany by Ralph Andrzejak [32]. The data consists of five groups, free EEG signals both in normal subjects and epileptic patients. The first two groups are recorded from five healthy subjects: with open (A) and closed eyes (B). The third and fourth groups are recorded prior to a seizure from part of the brain with the epilepsy syndrome (C) and from the opposite (healthy) hemisphere of the brain (D). The fifth group (E) is recorded from part of the brain with the epilepsy syndrome during the seizure. Three sets denoted A, C and E is used in this work. Each set contains 100 single channel EEG segments of 23.6-sec duration at a sampling rate of $f_s$ = 173.61 Hz. Set A consisted of segments taken from surface EEG recordings that were obtained from five healthy volunteers using a standardized electrode placement. Set E only contained seizure activity.

The object of wavelet analysis is to decompose signals into several frequency bands. Selection of appropriate wavelet and the number of decomposition levels are very important for the analysis of signals using DWT. The number of decomposition levels is chosen based on the dominant frequency components of the signal. The levels are chosen such that those parts of the signal that correlate well with the frequencies necessary for classification of the signal are retained in the wavelet coefficients. In this work, Daubechies 4 (db4) is selected because its smoothing feature was suitable for detecting changes of the EEG signals. Daubechies wavelets are the most popular wavelets representing foundations of wavelet signal processing, and are used in numerous applications. A detailed discussion about the characteristics of these wavelet functions can be found in the reference [14].

The frequency band $[f_m/2 : f_m]$ of each detail scale of the DWT is directly related to the sampling rate of the original signal, which is given by $f_m = f_s/2^{l+1}$, where $f_s$ is the sampling frequency, and $l$ is the level of decomposition. In this study, the sampling frequency is 173.6 Hz of the EEG signal. The highest frequency that the signal could contain, from Nyquist' theorem, would be $f_s/2$. Frequency bands corresponding to five decomposition levels for wavelet db4 with sampling frequency of 173.6 Hz of EEG signals were listed in Table I. The signals were decomposed into details D1-D5 and one final approximation A5.

TABLE I
FREQUENCY BANDS CORRESPONDING TO DIFFERENT DECOMPOSITION LEVELS

| DECOMPOSED SIGNALS | FREQUENCY BANDS (Hz) | DECOMPOSITION LEVEL |
|---|---|---|
| D1 | 43.4-86.8 | 1 (noises) |
| D2 | 21.7-43.4 | 2 (gama) |
| D3 | 10.8-21.7 | 3 (beta) |
| D4 | 5.40-10.8 | 4 (alpha) |
| D5 | 2.70-5.40 | 5 (theta) |
| A5 | 0.00-2.70 | 5 (delta) |

Classification of EEG signals requires the use of pattern recognition techniques. Pattern recognition is a process of perceiving a pattern of a given object based on the knowledge already possessed [33]. So automated pattern recognition uses various artificial intelligence techniques like fuzzy logic (FL), artificial neural networks (ANN) and adaptive fuzzy logic (AFL) for the classification of disturbance signals. Recently, techniques based on probabilistic models like Hidden Markov models, Dynamic time wrapping, Dempster-Shafer theory of evidence are also proposed.

An algorithm block diagram for classification of EEG signals is presented on Fig. 1. The algorithm structure is based on two stages: feature extraction stage (FES) and classification stage (CS). The input of the CS is a preprocessed signal. In this case, EEG signal in the time domain is transformed into the wavelet domain before applying as input to the CS. Feature extraction is the key for pattern recognition. A feature extractor should reduce the pattern vector (i.e., the original waveform) to a lower dimension, which contains most of the useful information from the original vector. In this algorithm, after realizing the FES (preprocessing), using detail and approximation coefficients in each decomposition level obtained from WT and MRA, the CS (processing) is implemented by using neural network (NN). NN are good at tasks such as pattern-matching and classification.

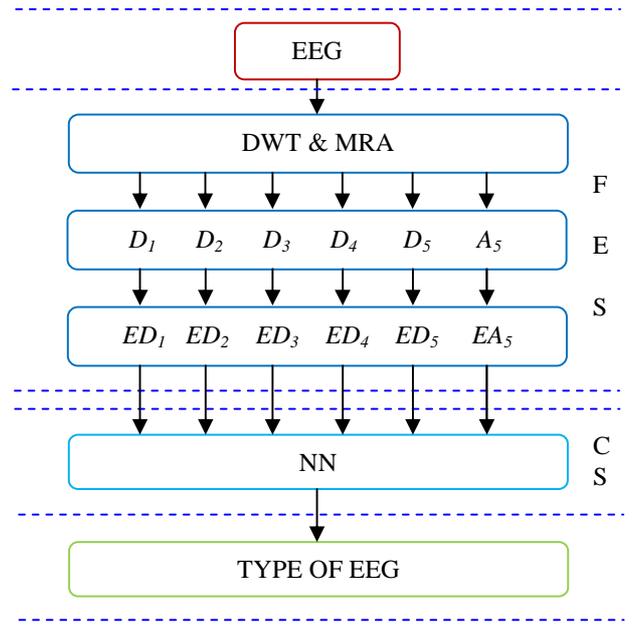

Fig. 1. Block diagram for classification of EEG signals

In the classification stage, the proposed wavelet energy distribution features are applied as input to NN. NN is a powerful pattern recognition tool. It is defined as software algorithms that can be trained to learn the relationships that exist between input and output data, including nonlinear relationships. Feed-Forward Neural Network (FFNN) is used to classify different EEG signals. The basic unit of a NN is the neuron, which realizes a function of weighted summation. A FFNN structure can be considered as an algebraic operator, such as weighted summation and multiplication. So, it is possible to reconstruct a wide class of algorithms by using a multiplier module.

III. ENERGY DISTRIBUTION OF EEG SIGNALS AND CLASSIFICATION RESULTS

Using MRA and Db4 wavelet function the three sets of the EEG signals (A set-100 EEG signals of the healthy patient, C set-100 EEG signals of the epilepsy patient in steady state and E set-100 EEG signals of the epilepsy patient during the seizure) was performed according to the percentage energy distribution of decomposition levels.

Energy distribution diagrams of EEG signals for different analyses cases are shown in Fig. 2. It shows the energy distribution of the analyzed signals: a) 100 samples of EEG signals of healthy subjects (A set); b) 100 samples of EEG signals of patients with epileptic syndrome (C set) and c) 100 samples of EEG signals of patients with epilepsy syndrome during the seizure (E set). It can be recognized different distribution of energy of the analyzed signals, which is generally quite similar for each group of EEG signals. It was noted in the EEG signal of healthy subjects, that energy activity in the frequency components of the D3 and D4 (beta and alpha) wave is quite similar and their percentage of the value of the total energy of the signal is around 20%. Energy activity in the frequency range D5 component (theta wave) is slightly lower of intensity and the percentage of its value in the total energy signal value is around 10%. Percentage value of total energy in the EEG signals in the frequency components of the D2 (gamma wave) is approximately 5%. Noise is negligibly small (D1) while the value of the frequency components of the D5 is about 45%, although for some samples it has a much higher value.

Unlike the distribution of EEG signals of healthy subjects, the energy distribution of the signal of patients with epilepsy syndrome is obviously different. In comparison with EEG signals of healthy subjects, components of D2, D3 and D4 of the EEG signals, the total energy distribution of EEG signals involved with a significantly lower percentage, while the values of D5 and A5 signal is much larger. Energy distribution of EEG signals in which it is registered epileptic seizure is significantly different from the first two cases. Energy activity component D3, D4 and D5 are dominant, while somewhat less value are the components of delta waves (A5).

The percentage of energy distribution can be used for classification of EEG signals.

NN are highly interconnected simple processing units designed in a way to model how the human brain performs a particular task. Each of those units, also called neurons, forms a weighted sum of its inputs, to which a constant term called bias is added. This sum is then passed through a transfer function: linear, sigmoid or hyperbolic tangent. The choice of number of hidden layers and the number of neurons in each layer is one of the most critical problems in the construction of neural architecture. In order to find the optimal network architecture, several combinations should be evaluated. These combinations include networks with different number of hidden layers, different number of units in each layer and different types of transfer functions. The FFNN model was provided in Matlab.

Based on the feature extraction, 6-dimensional feature sets (D1, D2, D3, D4, D5 and A5) for training and testing data were constructed. The dimensions here describe different features resulting from the wavelet transform, that is to say, the total size of training data or testing data set is 6×300. Considering the classification performance of this method, this input vector is applied as the input to the WNN structure. The training parameters and the structure of the WNN used in this study are shown in Table II.

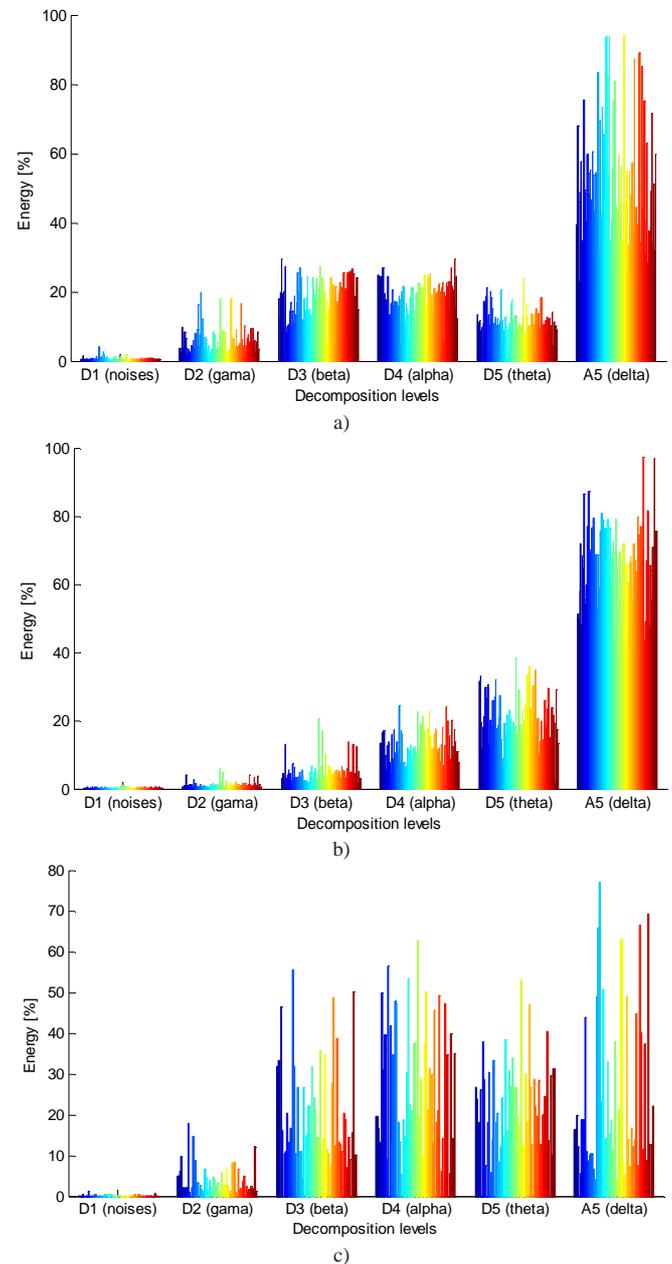

Fig. 2. Energy distribution diagram (%) of the a) A set-100 EEG signals of the healthy patient, b) C set-100 EEG signals of the epilepsy patient in steady state and c) E set-100 EEG signals of the epilepsy patient during the seizure

They were selected to obtain best performance, after several different experiments, such as the number of hidden layers, the size of the hidden layers, value of the moment constant and learning rate, and type of the activation functions. The data for each experiment are selected randomly. In Table III presented are classification results of WNN algorithm where 250 data sets were used to train the NN model and 50 data sets were used for testing process. The system can correctly classify 47 of the 50 different EEG signals in the testing set, as shown in Table III. The classified accuracy rate of the EEG signals of the proposed approach was 94.0%. Hundred percent correct classification rates are obtained for normal EEG signals.

TABLE II
NN ARCHITECTURE AND TRAINING PARAMETERS

| ARCHITECTURE | |
|---|---|
| THE NUMBER OF LAYERS | 3 |
| THE NUMBER OF NEURON ON THE LAYERS | INPUT: 6, HIDDEN: 5, OUTPUT: 1 |
| THE INITIAL WEIGHTS AND BIASES | RANDOM |
| ACTIVATION FUNCTIONS | TANGENT SIGMOID |
| TRAINING PARAMETERS | |
| LEARNING RULE | LEVENBERG–MARQUARDT BACK-PROPAGATION |
| MEAN-SQUARED ERROR | 1E-01 |

TABLE III
EEG CLASSIFICATION RESULTS OF WNN ALGORITHM

| CLASS | HEALTHY | EPILEPSY SYNDROME | SEIZURE | ACCURACY [%] |
|---|---|---|---|---|
| HEALTHY | 16 | 0 | 0 | 100.0 |
| EPILEPSY SYNDROME | 2 | 17 | 0 | 88.2 |
| SEIZURE | 0 | 1 | 14 | 92.9 |
| OVERALL SUCCESS RATE | | | | 94.0 |

IV. CONCLUSION

Epileptic seizures are manifestations of epilepsy. The detection of epileptiform discharges in the EEG is an important component in the diagnosis of epilepsy. As EEG signals are non-stationary, the conventional method of frequency analysis is not highly successful in diagnostic classification. In this paper, an algorithm for classification of EEG signal based on WT and PRT has been proposed. DWT with the MRA is applied to decompose EEG signal at resolution levels of the components of the EEG signal and to extract the percentage distribution of energy features of the EEG signal at different resolution levels. The FFNN classifies these extracted features to identify the EEGs type according to the percentage distribution of energy features. The results showed that the proposed classifier has the ability of recognizing and classifying EEG signals efficiently. The most important advantage of the proposed method is the reduction of data size as well indicating and recognizing the main characteristics of signal. Furthermore, it can reduce memory space, shorten pre-processing needs, the network size and increase computation speed for the classification of an EEG signal.